\documentclass[11pt,a4paper]{article}
\usepackage[hyperref]{emnlp2020}
\usepackage{times}
\usepackage{latexsym}

\usepackage{microtype}
\usepackage{multirow}
\usepackage{csvsimple}
\usepackage{hyperref}
\usepackage{graphicx}
\usepackage{amssymb}
\usepackage{enumitem,amssymb}
\newlist{todolist}{itemize}{2}
\setlist[todolist]{label=$\square$}
\usepackage{pifont}
\newcommand{\cmark}{\ding{51}}%
\newcommand{\done}{\rlap{$\square$}{\raisebox{2pt}{\large\hspace{1pt}\cmark}}%
\hspace{-2.5pt}}

\newcounter{notecounter}
\newcommand{\enotesoff}{\long\gdef\enote##1##2{}}
\newcommand{\enoteson}{\long\gdef\enote##1##2{{
\stepcounter{notecounter}
{\large\bf
\hspace{1cm}\arabic{notecounter} $<<<$ ##1: ##2
$>>>$\hspace{1cm}}}}}
\enoteson
\enotesoff

\aclfinalcopy %
\def\aclpaperid{2513} %

\def\figref#1{Figure~\ref{fig:#1}}
\def\figlabel#1{\label{fig:#1}\label{p:#1}}

\def\tabref#1{Table~\ref{tab:#1}}
\def\tablabel#1{\label{tab:#1}\label{p:#1}}

\def\eqref#1{Eq.~\ref{eqn:#1}}

\title{BERT-kNN:
Adding a kNN Search Component to Pretrained Language Models
for Better QA}

\author{Nora Kassner,  Hinrich Sch\"utze \\
  Center for Information and Language Processing (CIS) \\
  LMU Munich, Germany \\
  \texttt{kassner@cis.lmu.de}}
\date{}

\long\def\eat#1{\ignorespaces}

\begin{document}
\maketitle
\begin{abstract}

  \citet{khandelwal20generalization} use
a k-nearest-neighbor (kNN) component to improve language model
performance.
We show that
this idea is beneficial for
open-domain question answering (QA).
To improve the 
recall of facts encountered during training,
we combine BERT
\cite{devlin-etal-2019-bert} with a traditional information retrieval step (IR) and a
kNN search over a large datastore of an embedded text collection.
Our contributions are as follows: i) BERT-kNN outperforms BERT on cloze-style QA by large margins without any
further training. ii) We show that BERT often identifies the
correct response category (e.g.,
US city),
but only kNN recovers the
factually correct answer (e.g.,
``Miami'').
iii) Compared
to BERT, BERT-kNN
excels for rare facts. iv) BERT-kNN can  easily handle
facts not covered by BERT's training set, e.g., recent events.

\end{abstract}

\section{Introduction}
Pretrained language models (PLMs) like BERT
\cite{devlin-etal-2019-bert}, GPT-2 \cite{radford2019language} and RoBERTa \cite{DBLP:journals/corr/abs-1907-11692} have emerged as universal tools
that not only capture a diverse range of linguistic, but also
(as
recent evidence seems to suggest) factual knowledge.

\citet{petroni2019language}
introduced LAMA (LAnguage Model Analysis) to
test BERT's performance on open-domain QA
and therefore investigate
PLMs'  capacity to
recall factual knowledge without the use of finetuning.
Since the  PLM  training
objective is to predict masked tokens,
question answering tasks
can be reformulated as cloze questions; e.g., ``Who
wrote `Ulysses'?''
is reformulated as ``[MASK] wrote `Ulysses'.'' In this setup,
\citet{petroni2019language} show that, on QA,  PLMs
outperform baselines trained on
automatically extracted knowledge
bases (KBs).

\begin{table*}
\small
\centering
\begin{tabular}{lllllll}
Dataset &BERT-base& BERT-large&ERNIE&Know-BERT&E-BERT&BERT-kNN \\\hline
LAMA &27.7&30.6&30.4&31.7&36.2&\textbf{39.4}\\
LAMA-UHN &20.6&23.0&24.7&24.6&31.1&\textbf{34.8}
\end{tabular}
\caption{\tablabel{UHN_rare} Mean P@1 on LAMA and LAMA-UHN
  on the TREx and GoogleRE subsets  for BERT-base,
  BERT-large, ERNIE
  \cite{zhang2019ernie}, KnowBert
  \cite{Peters2019KnowledgeEC}, E-BERT
  \cite{Poerner2019BERTIN} and BERT-kNN. BERT-kNN performs best.}
\end{table*}

Still, given that PLMs have seen more text than 
humans  read in a lifetime, their performance on open-domain QA seems poor.
Also, many LAMA facts that PLMs do get right are not 
``recalled'' from  training, but are guesses instead
\cite{Poerner2019BERTIN}.
To address PLMs' poor performance on facts and
choosing BERT as our PLM,
we  introduce BERT-kNN.
\enote{nk}{BERT-kNN combines BERT's predictions with a
kNN search over a text collection where the text collection can be BERT's
training set or any other suitable text corpus.
Due to its kNN component and its resulting ability to
directly access facts stated in the searched text,
BERT-kNN outperforms BERT
on cloze-style QA by large
margins.}

\begin{figure}
  \includegraphics[width=\linewidth]{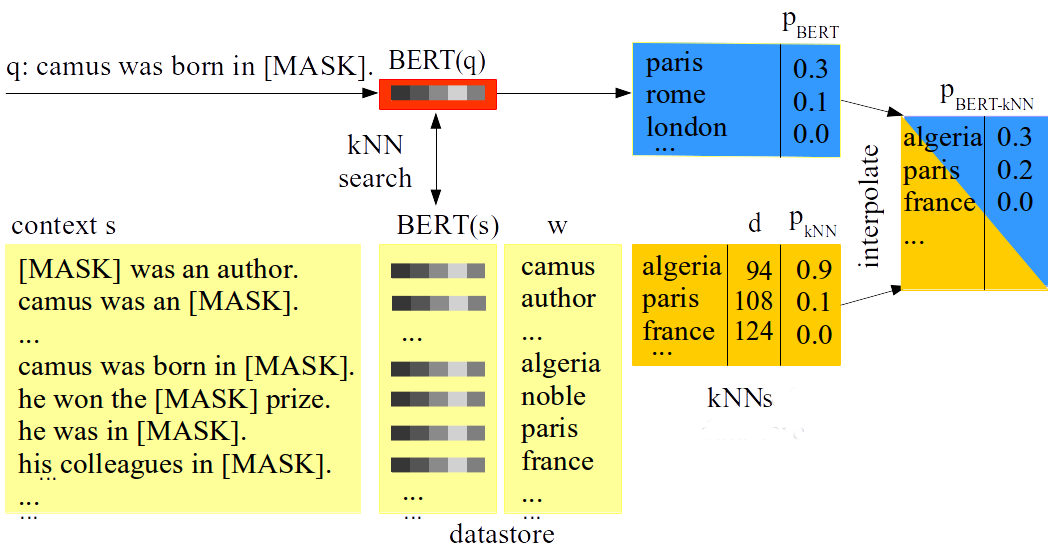}
  \caption{\figlabel{model} BERT-kNN interpolates BERT's
    prediction for question $q$ with a kNN-search.
    The kNN search runs in BERT's embedding space, comparing the embedding of $q$ with the embeddings of a retrieved subset of a large text collection:
    Pairs of
a word $w$ in the text collection and the BERT embedding of $w$'s
    context ($BERT(s)$) 
are stored in a key-value datastore. An IR step is used to define a relevant subset of the full datastore (yellow).
$BERT(q)$ (red) is  BERT's embedding of the question.
The kNN search runs between $BERT(q)$ and $BERT(s)$ and the corresponding
distance $d$ and word $w$
is returned (orange).
Finally, BERT's predictions (blue)
are interpolated with this kNN search result.
 }
\end{figure}

BERT-kNN combines BERT's predictions with a
kNN search. The kNN search runs in BERT's embedding space, comparing the embedding of the question with the embeddings of a retrieved subset of a large text collection.
The text collection can be BERT's
training set or any other suitable text corpus.
Due to its kNN component and its resulting ability to
directly access facts stated in the searched text,
BERT-kNN outperforms BERT
on cloze-style QA by large
margins.

A schematic depiction of the model is shown in \figref{model}. Specifically, we use BERT
to embed each token's masked context $s$ in the text collection ($BERT(s)$).
Each pair of context embedding and token
is stored as a key-value pair in a datastore.
Testing for a cloze question $q$,
the embedding of $q$ ($BERT(q)$) serves as query  to find the $k$ context-target pairs in the subset of the datastore that are closest.
The final prediction is an
interpolation of the kNN search and the PLM predictions. 
\enote{nk}{To make this more effective, we
first query a separate information retrieval (IR) index with
the original question $q$ and 
only search over the top hits when finding the $k$ nearest
neighbors of $BERT(q)$ in embedding space.
}

We find that the kNN search over the full datstore alone does not obtain good results. Therefore, we
first query a separate information retrieval (IR) index with
the original question $q$ and 
only search over the most relevant subset of the full datastore when finding the $k$-nearest-neighbors of $BERT(q)$ in embedding space.

We
find that the PLM often correctly predicts the answer category
and therefore the correct answer is often among the top $k$-nearest-neighbors. A typical example is 
``Albert Einstein was born in [MASK]'': the PLM knows that a city
is likely to follow and maybe even that it is a German city, but
it fails to pick the correct city.
On the other hand, the top-ranked answer in the kNN search
is ``Ulm'' and so the correct filler for the mask can be identified.

BERT-kNN
sets a new state-of-the-art on the LAMA cloze-style QA
dataset without any further training. Even though BERT-kNN
is based on BERT-base, it also outperforms BERT-large. 
The performance gap between BERT and
BERT-kNN is most pronounced on hard-to-guess facts.
Our method can also make recent events available to BERT
without any need of retraining: we can
simply add embedded text collections covering recent events to 
BERT-kNN's datastore.

The source code of our experiments is available under:
\url{https://github.com/norakassner/BERT-kNN}.

\section{Data}
The LAMA dataset is a cloze-style QA dataset that allows to
query PLMs for facts in a way analogous to KB queries.  A cloze question
is generated using a subject-relation-object triple from a
KB and a templatic statement for the
relation that contains variables X and Y for subject and
object; e.g, ``X was born in Y''.  The subject is
substituted for X and [MASK] for Y.
In all LAMA triples, 
Y is  a single-token answer.

LAMA covers different sources:
The
GoogleRE\footnote{\url{https://code.google.com/archive/p/relation-extraction-corpus/}}
set covers the relations ``place of birth'',
``date of birth'' and ``place of death''.
TREx
\cite{DBLP:conf/lrec/ElSaharVRGHLS18} consists of a subset
of Wikidata triples covering 41 relations.
ConceptNet
\cite{li-etal-2016-commonsense} combines 16 commonsense
relations among words and phrases. The underlying
Open Mind Common Sense corpus provides matching statements
to query the language model. SQuAD
\cite{rajpurkar-etal-2016-squad} is a standard question
answering dataset. LAMA contains a subset of 305
context-insensitive questions.
Unlike KB queries, SQuAD uses manually reformulated cloze-style questions which are not based on a template.

We use  SQuAD and an additional 305 ConceptNet queries for hyperparamter search.

\citet{Poerner2019BERTIN} introduce LAMA-UHN, a subset of
LAMA's TREx and GoogleRE questions
from which easy-to-guess facts have been removed.

To test BERT-kNN's performance on unseen facts, we collect Wikidata triples containing TREx relations from Wikipedia pages created  January--May 2020 and add them to the datastore.

\eat{

\begin{table}
\small
\centering
\begin{tabular}{l|l}
  Configuration &P@1 \\\hline
1st hidden layer &36.8\\
2nd last hidden layer &\textbf{39.4}\\
3rd hidden layer &34.7\\
mean 1st hidden layer&27.7\\
without IR, 2nd hidden layer & 26.9 
\end{tabular}
\caption{\tablabel{embeddings} Mean P@1 for LAMA on the TREx
  and GoogleRE subsets for different embedding configurations.}
\end{table}

}

\begin{table}
\small
\centering
\begin{tabular}{lrrrrr}
\hline \textbf{Dataset} & \multicolumn{2}{c}{\textbf{Statistics}} & \multicolumn{3}{c}{\textbf{Model}}\\ \hline
 & \multirow{2}{*}{ Facts} & \multirow{2}{*}{Rel} &\multirow{2}{*}{ BERT} & \multirow{2}{*}{kNN} &BERT \\
  & & & & &-kNN \\\hline
GoogleRE  & 5527 & 3 & 9.8  &\textbf{51.1}& 48.6\\
TREx &  34039& 42  &29.1 &34.4&\textbf{38.7}  \\
ConceptNet  &  11153& 16&\textbf{15.6}&4.7&11.6\\ \hline
SQuAD  & 305 & - & 14.1 & \textbf{25.5}&24.9\\\hline
unseen  &  34637& 32&18.8 & 21.5&\textbf{27.1}\\
\hline
\end{tabular}
\caption{\tablabel{results} Mean P@1 for BERT-base, kNN
  and their interpolation  (BERT-kNN) for LAMA subsets and 
  unseen facts.
BERT results differ from \citet{petroni2019language} where a  smaller vocabulary is used.}
\end{table}

\begin{table}
\small
\centering
\begin{tabular}{l|l}
  Configuration &P@1 \\\hline
 hidden layer 12 &36.8\\
 hidden layer 11 &\textbf{39.4}\\
 hidden layer 10 &34.7\\\hline
hidden layer 11 (without IR) & 26.9 
\end{tabular}
\caption{\tablabel{embeddings} Mean P@1 on LAMA (TREx, GoogleRE subsets) for different context embedding strategies. Top: The context embedding is represented by the embedding of the masked token in different hidden layers.
Best performance is obtained using BERT's hidden layer 11. Bottom: We show that BERT-kNN's performance without the additional IR step drops significantly. We therefore conclude that the IR step is an essential part of BERT-kNN.}
\end{table}

\section{Method}
BERT-kNN combines BERT with a kNN search component. 
Our  method is generally applicable to  PLMs.
Here, we use BERT-base-uncased \cite{devlin-etal-2019-bert}.
BERT is pretrained on the BookCorpus  \cite{Zhu_2015_ICCV} 
and the English Wikipedia.

\textbf{Datastore.}
Our text collection $C$ is the 2016-12-21 English Wikipedia.\footnote{\url{dumps.wikimedia.org/enwiki}}
For each single-token word occurrence $w$ in a sentence in $C$, 
we compute the pair $(c,w)$ where $c$ is a context
embedding computed by BERT.
To be specific, we mask the occurrence of $w$ in the sentence and use the embedding of the masked
token.
We store all pairs $(c,w)$ in a key-value datastore $D$ where
$c$ serves as key and $w$ as value.

\textbf{Information Retrieval.} We find that just using the
datastore $D$ does not give good results (see result section).
We therefore
use \citet{Chen2017ReadingWT}'s IR system to
first select a small subset of $D$ using a
keyword search. The IR index contains all Wikipedia
articles. An article is represented as a bag of words and
word bigrams. We find
the top 3 relevant Wikipedia articles using TF-IDF search. 
For KB queries, we use the subject to query the IR index.
If the subject has its dedicated Wikipedia page, we simply use this.
For non-knowledge base queries, we use the cloze-style question $q$ 
([MASK] is removed).

\textbf{Inference.}
During testing,
we first run the IR search to identify
the subset $D'$ of $D$ that corresponds to the relevant
Wikipedia articles.
For the kNN search, $q$ is embedded in the same way as the context representations $c$ in $D$:
we set $BERT(q)$ to the embedding
computed by BERT for [MASK].
We then retrieve the $k=128$ nearest-neighbors of $BERT(q)$ in
$D'$.
We convert the distances (Euclidean) between $BERT(q)$ and the kNNs to a probability distribution using softmax.
Since a word $w$ can occur several times in kNN, we compute its final output
probability as the sum over all occurrences.

In the final step,
we interpolate 
kNN's (weight 0.3)
and BERT's original predictions (weight 0.7).
We optimize hyperparameters on  dev.
See supplementary for details.

\begin{figure}
  \includegraphics[width=0.9\linewidth]{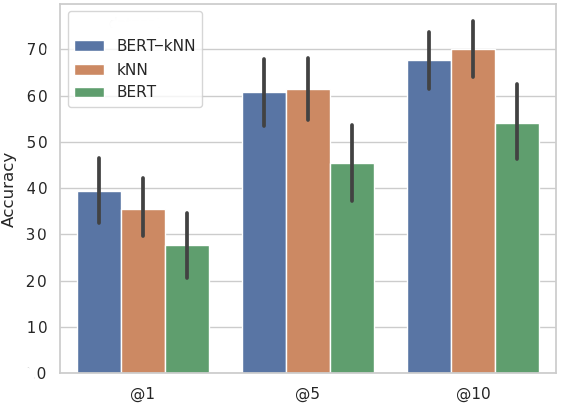}
  \caption{ \figlabel{1_5_10} Mean P@1, P@5, P@10 on LAMA for original BERT and BERT-kNN.}

\end{figure}

\textbf{Evaluation.}
Following \citet{petroni2019language} we report mean
precision at rank $r$ (P@r). P@r is 1
if the top $r$ predictions contain the correct answer, otherwise it returns 0.
To compute mean precision, we first average within each relation and then across relations.

\section{Results and Discussion}
\tabref{UHN_rare} shows that BERT-kNN outperforms BERT on
LAMA.  It has about 10 precision point gain over BERT,
base and large. Recall that BERT-kNN uses BERT-base.  The
performance gap between original BERT and BERT-kNN becomes
even larger when evaluating on LAMA-UHN, a subset of LAMA
with hard-to-guess facts.

It also outperforms entity-enhanced versions of BERT (see
related work) -- ERNIE \cite{zhang2019ernie}, KnowBert
\cite{Peters2019KnowledgeEC} and E-BERT
\cite{Poerner2019BERTIN} -- on LAMA.

\begin{table*}
\small
\centering
\tabcolsep=0.11cm
\begin{tabular}{lll}
\hline  &\textbf{Query and True Answer}& \textbf{Generation} \\ \hline
\multirow{
  3}{*}{\rotatebox[origin=c]{90}{\scriptsize \ \ \begin{tabular}{c}Google
      \\ RE\end{tabular}}}
&hans gefors was born in [MASK].  &BERT-kNN: stockholm (0.36), oslo (0.15), copenhagen (0.13)\\
 &True: stockholm& BERT: oslo (0.22), copenhagen (0.18), bergen (0.09) \\
 & & kNN: stockholm (1.0), lund (0.00), hans (0.00)\\\cline{2-3}
\multirow{
  3}{*}{\rotatebox[origin=c]{90}{\scriptsize \ \ \begin{tabular}{c}TREx\end{tabular}}}
&regiomontanus works in the field of [MASK].  & BERT-kNN: astronomy (0.20), mathematics (0.13), medicine (0.06)\\
 &True: mathematics  &BERT: medicine (0.09), law (0.05), physics (0.03)\\
 && kNN: astronomy (0.63), mathematics (0.36), astronomical (0.00) \\\cline{2-3}
\multirow{
 3}{*}{\rotatebox[origin=c]{90}{\scriptsize \ \ \begin{tabular}{c}Concept\\Net\end{tabular}}}
& ears can [MASK] sound. & BERT-kNN: hear (0.27), detect (0.23), produce (0.06)\\
&True: hear & BERT: hear (0.28), detect (0.06), produce (0.04)\\
&  & kNN: detect (0.77), hear (0.14), produce (0.10) \\\cline{2-3}
\multirow{
  3}{*}{\rotatebox[origin=c]{90}{\scriptsize \ \ \begin{tabular}{c}Squad\end{tabular}}}
 &tesla was in favour of the [MASK] current type.& BERT-kNN: alternating (0.39), electric (0.18), direct (0.11) \\
&True: ac &  BERT: electric (0.28), alternating (0.18), direct (0.11)\\
&  &  kNN: alternating (0.87), direct (0.12), ac (0.00) \\\cline{2-3}
\end{tabular}
\caption{\tablabel{examples} Examples of generation for BERT-base, kNN, BERT-kNN. The last column reports the top three tokens generated together with the associated probability (in parentheses).}
\end{table*}

\tabref{results} shows  that BERT-kNN outperforms BERT on 3
out of 4 LAMA subsets.
BERT prevails on ConceptNet; see discussion below.
Huge gains are obtained on the GoogleRE dataset. %
\figref{1_5_10} shows precision at 1, 5 and 10. BERT-kNN
performs better than BERT in all three categories. 

\tabref{embeddings} compares different context embedding
strategies.  BERT's masked token embedding of hidden layer
11 performs best. We also show the necessity of the IR step by running a kNN search over all Wikipedia contexts, which results in precision lower than original BERT.
To run an efficient kNN search over all contexts instead of the relevant subset identified by the IR step, we use the FAISS libary  \cite{JDH17}.

\tabref{results} also shows that neither BERT nor kNN
alone are sufficient for top performance, while the
interpolation of the two yields optimal results. In many
cases, BERT and kNN are complementary. kNN is worse than
BERT on ConceptNet, presumably because commonsense knowledge
like ``birds can fly'' is less well-represented in Wikipedia
than entity triples and also because relevant articles are harder
to find by IR search.
We keep the
interpolation parameter constant over all
datasets.
\tabref{examples}
shows that kNN often has high confidence for correct answers
-- in such cases it is likely to dominate  less
confident predictions by BERT. The converse is also true
(not shown).
Further optimization could be obtained by
tuning interpolation per dataset.

\enote{hs}{ i don't understand this sentence:

BERT-kNN is inherently able to distinguish when to
rely more on BERT or the kNN predictions.

if interpolation is fixed, then isn't also the reliance on
one source vs the other fixed?

}

BERT-kNN  answers facts unseen during
pretraining better than BERT, see \tabref{results}.
BERT was not trained on 2020 events, so it must resort to guessing.
Generally, we see that BERT's knowledge is mainly based on guessing as it has seen Wikipedia during training but is not able to recall the knowledge recovered by kNN.

\tabref{examples} gives examples for BERT and
BERT-kNN predictions. We see that  BERT predicts the answer category correctly,
but it often needs help from kNN  to recover the
correct entity within that category.

\section{Related work}
PLMs are top performers for many tasks, including QA
\cite{qa,Alberti2019ABB, DBLP:journals/corr/abs-1906-05317}.  \citet{petroni2019language}
introduced the LAMA QA task to probe PLMs'
knowledge of facts typically modeled by KBs.

The basic idea of BERT-kNN is similar to 
\citet{khandelwal20generalization}'s interpolation of
a PLM and kNN for language modeling. In 
contrast, we  address QA.  We
introduce an IR step into the model that is essential for
good performance.
Also, our context representations differ as we use embeddings of the
masked token.

\citet{Grave2016ImprovingNL} and
\citet{DBLP:journals/corr/MerityXBS16}, inter alia,
also make use of memory to store
hidden states.
 They focus on recent
 history, making it easier to copy rare vocabulary items.
 
DRQA \cite{Chen2017ReadingWT} is an open-domain QA
model that combines an IR step with a neural reading
comprehension model. We use the same IR module, but
our model differs significantly.  DRQA does not predict
masked tokens, but extracts answers from text. It does not 
use PLMs nor a kNN module. Most importantly,
BERT-kNN is fully unsupervised and does not require
any extra training.

Some work on knowledge in PLMs focuses on injecting
knowledge into BERT's encoder. ERNIE \cite{zhang2019ernie}
and KnowBert \cite{Peters2019KnowledgeEC} are
entity-enhanced versions of BERT. They introduce additional
encoder layers that are integrated into BERT's original
encoder by expensive additional pretraining. \citet{Poerner2019BERTIN}
injects factual entity knowledge into BERT's embeddings
without pretraining  by aligning Wikipedia2Vec
entity vectors \cite{DBLP:journals/corr/abs-1812-06280} with
BERT's wordpiece vocabulary. This approach is also limited
to labeled entities. Our approach on
the other hand is not limited to labeled entities nor does
it require any  pretraining. Our approach is conceptually 
different from entity-enhanced versions of BERT and could potentially
be combined with them for even better performance.
Also, these models address language modeling, not QA.

The combination of PLMs with an IR step/kNN search has attracted a lot of recent research interest. The following paragraph lists concurrent work: 

\citet{petroni2020context} also combine BERT with an IR step to improve cloze-style QA. They do not use a kNN search nor an interpolation step but feed the retrieved contexts into BERT's encoder.
\citet{guu2020realm} augment PLMs with a latent knowledge retriever. In contrast to our work they continue the pretraining stage. They jointly optimize the masked language modeling objective and backpropagate through the retrieval step.
\citet{lewis2020retrievalaugmented, izacard2020leveraging} leverage retrieved contexts for better QA using finetuned generative models. They differ in that the latter fuse evidence of multiple contexts in the decoder. 
\citet{joshi2020contextualized} integrate retrieved contexts into PMLs for better reading comprehension.

\section{Conclusion}
This work introduced BERT-kNN, an interpolation of BERT
predictions with a kNN search for unsupervised cloze-style
QA. BERT-kNN sets a state-of-the-art on  LAMA
without any further
training. BERT-kNN
can be easily enhanced with knowledge about new events that
are not covered in the training text used for pretraining BERT.

In future work, we want to exploit the utility of the kNN
component for explainability: kNN predictions are based on
retrieved contexts, which can be shown to users to justify
an answer.

\section{Acknowledgements}

This work has been funded by the German Federal Ministry of Education
and Research (BMBF) under Grant No. 01IS18036A. The authors of this work take full responsibility for its content.

\bibliography{anthology}
\bibliographystyle{acl_natbib}

\newpage

\cleardoublepage

\appendix
\section{Data}
\label{app_unseen}
LAMA and LAMA-UHN can be downloaded from:
\url{ https://dl.fbaipublicfiles.com/LAMA/}

For TREx unseen, we downloaded the latest Wikidata and  Wikipedia dump from:\\
\url{https://dumps.wikimedia.org/wikidatawiki/entities/wikipedia_en/latest-all.json.bz2}\\
and \\
\url{https://dumps.wikimedia.org/enwiki/latest/enwiki-latest-pages-articles.xml.bz2}.

We filter for TREx relations and only consider facts which have a Wikipedia page created after January 1st 2020. We only consider relations with 5 questions or more. We add the additional embedded Wikipedia articles to the datastore.

\section{Inference}
\label{math}
The probability of the kNN search for word $w$ is given by:\\
$p_{kNN}(w \mid q) \sim \sum_{(c_{w}, w) \in kNN} e^{-d(BERT(q),c_{w})/l}$.\\
\\
The final probability of BERT-kNN is the interpolation of the predictions of BERT and the kNN search:\\
$p_{{\tiny BERT-kNN}}(q) $
= $\lambda p_{{\tiny kNN}}(q)  +(1-\lambda) p_{{\tiny BERT}}(q)$,\\
\\
with\\
$q$ question,\\
$BERT(q)$ embedding q,\\
$w$ target word,\\
$s_{w}$ context of w,\\
$c_{w}= BERT(s)$ embedded context,\\
$d$ distance,\\
$l$ distance scaling,\\
$\lambda$ interpolation parameter.

\section{Hyperparameters}
\label{app_hyper}
Hyperparameter optimization is done with the 305 SQuAD questions and additional randomly sampled 305 ConceptNet questions. We remove the 305 ConceptNet questions from the test set. We run the hyperparameter search once.\\
We run a grid search for the following hyperparameters: \\
    Number of documents $N$ = $ [ $1, 2, 3, 4, 5$]$,\\
    Interpolation $\lambda$ = $ [ $0.2, 0.3, 0.4, 0.5, 0.6, 0.7, 0.8$ ] $,\\
    Number of NN $k$ =  $ [ $64, 128, 512$ ] $,\\
    Distance scaling $l$ = $ [ $5, 6, 7, 8, 9, 10, 11, 12$ ] $.\\
  \\
  The optimal P@1 was found for:\\
    Number of documents $N$ = 3,\\
    Interpolation parameter $\lambda$ =  0.3,\\
    Number of NN $k$= 128,\\
    Distance scaling $l$ =  6.\\

\section{kNN without IR}
To enable a kNN search over the full datastore we use FAISS index \cite{JDH17}. We train the index using 1M randomly sampled keys and 40960 number of clusters. Embeddings are quantized to 64 bytes. During inference the index looks up 64 clusters.

\section{Computational Infrastructure}
\label{comp}

    The creation of the datastore is computationally expensive but only a single forward pass is needed. The datastore creation is run on a server with 128 GB memory,  Intel(R) Xeon(R) CPU E5-2630 v4, CPU rate 2.2GHz,  number of cores 40(20), 8x GeForce GTX 1080Ti.
    One GPU embedds 300 contexts/s. The datastore includes 900M contexts.\\
    \\
    Evaluation is run on a server with 128 GB memory, Intel(R) Xeon(R) CPU E5-2630 v4, CPU rate 2.2GHz, number of cores 40(20).
    Evaluation time for one query is 2 s but code can be optimized for better performance.

\end{document}